# Predicting the Transportation Activities of Construction Waste Hauling Trucks: An Input-Output Hidden Markov Approach

Hongtai Yang[a], Boyi Lei[a], Ke Han[a,*], Luna Liu[a]

[a]*School of Transportation and Logistics, National Engineering Laboratory of Integrated Transportation Big Data Application Technology, National United Engineering Laboratory of Integrated and Intelligent Transportation, Institute of System Science and Engineering, Southwest Jiaotong University, Chengdu, China, 611756*

**Abstract**

Construction waste hauling trucks (CWHTs), as one of the most commonly seen heavy-duty vehicles in major cities around the globe, are usually subject to a series of regulations and spatial-temporal access restrictions because they not only produce significant NOx and PM emissions but also causes on-road fugitive dust. The timely and accurate prediction of CWHTs' destinations and dwell times play a key role in effective environmental management. To address this challenge, we propose a prediction method based on an interpretable activity-based model, input-output hidden Markov model (IOHMM), and validate it on 300 CWHTs in Chengdu, China. Contextual factors are considered in the model to improve its prediction power. Results show that the IOHMM outperforms several baseline models, including Markov chains, linear regression, and long short-term memory. Factors influencing the predictability of CWHTs' transportation activities are also explored using linear regression models. Results suggest the proposed model holds promise in assisting authorities by predicting the upcoming transportation activities of CWHTs and administering intervention in a timely and effective manner.

*Keywords:* Construction Waste Hauling Trucks; Human Mobility; Machine Learning; Urban Transportation; LSTM

## 1. Introduction

With the development of information and communication technologies (ICT) comes pervasive monitoring of heavy-duty vehicles in cities for environmental and transport management. Such monitoring is typically achieved through GPS positioning (Arebey et al., 2011), license plate recognition (Huang et al., 2022), or radio frequency identification (Hannan et al., 2011). This paper focuses on construction waste hauling trucks (CWHTs), which is commonly seen in major cities around the globe. In China, CWHTs are within the purview of urban environmental management, as they not only produce significant NOx and PM emissions but also cause on-road and on-site fugitive dust.

Many Chinese cities have imposed access restrictions on CWHTs during Heavy Pollution Episodes to alleviate the accumulation of air pollutants. There are, however, instances of violation of such restrictions. Luckily, GPS-based CWHT monitoring enables the detection of area infringement by tracking vehicle locations in real time. However,

* Corresponding author.

*E-mail address:* yanghongtai@swjtu.cn (Hongtai Yang), 2018113572@my.swjtu.edu.cn (Boyi Lei), kehan@swjtu.edu.cn (Ke Han), l18935318480@163.com (Luna Liu)

following this, law enforcement units are required to engage these trucks on site to collect evidence and administer intervention, which is clearly challenged by time delays related to unit dispatch and data transmission[1]. Therefore, predicting the whereabouts of CWHTs is a crutial component of such an endeavor, and this paper aims to accomplish such a task.

The transportation activities of CWHTs encompass an array of intermediate locations (e.g. construction sites, waste disposal sites, refuel stations, parking spaces), each termed a stay point. This work is concerned with short-term prediction of the next stay point and dwell time for individual trucks. Most literature on mobility predictions based on stay points focused on human motilities (Wei et al., 2021; Wu et al., 2019), and researchers explored human travel behavior based on location information from mobile phone signal data (Do et al., 2015; Neto et al., 2015; Xiao et al., 2015; Yin et al., 2018), taxi trajectory data (Lv et al., 2018), shared bicycle trajectory data (Wang et al., 2022), location-based social networks data (Comito, 2020), etc. In this work, we consider the input-output hidden Markov model (IOHMM) via an in-depth analysis of the spatial-temporal information of truck trajectories, which enhances both prediction performance and model interpretability.

Before constructing the activity sequences of CWHTs, we implement a data processing method by extracting individual stay points from continuous trajectory records, followed by a grid-based refinement of location information within these points. Then, we cluster the transportation activities of CWHTs into groups, defined as distinctive transportation activity types, which are incorporated as hidden states within the IOHMM. Specifically, a CWHT's activity type varies based on its transportation task, exhibiting a correlation with both destination and dwell time. Leveraging such activity types, the proposed model not only achieves superior prediction performance over a few benchmarks (long short-term memory, Markov chain, linear regression), but also has good interpretability, by allowing regulators to understand the spatiotemporal characteristics of CWHT movements across different transportation activity types by examining the estimated parameters.

The rest of this paper is organized as follows. Section 2 provides a literature review concerning CWHT and related studies on travel prediction. Section 3 introduces the dataset and the methodological framework of the IOHMM used for predicting CWHT transportation activities. Following this, Section 4 demonstrate the model's predictive results and conducts an extensive performance analysis. Finally, Section 5 presents a summary of the primary work made in the paper, emphasizes the implications of the research, and explores future research directions.

## 2. Literature Review

Recently, the number of studies related to CWHTs has been growing, although still limited, due to concerns about the environmental impact of CWHTs. Lu (2019) used big data analytics to investigate the possible determinants of the illegal dumping behavior of CWHTs in Hong Kong using different transportation behaviors as indicators. Lu et al. (2022) also explored the loading patterns of CWHTs in Hong Kong, as well as the factors influencing these patterns, using a triangulation method based on big data and driver interviews. Wei et al. (2022) analyzed the relationship between CWHT transportation characteristics and carbon emissions based on construction waste transportation records in Hong Kong. These studies have provided evidence of pollution and environmental harm caused by CWHTs and support for the development of effective regulation policies. Other studies have focused on path optimization in various circumstances (Akegawa et al., 2022; Choi & Nieto, 2011; Choi et al., 2009; Sulemana et al., 2018) and the identification of hotspot areas in the transportation process (Bi et al., 2022). We discovered that thorough trajectory studies on CWHTs are still lacking. To fill this gap, this study makes an attempt to predict the location and time of arrival of the next destination of CWHT based on the historical trajectory of the truck.

---

[1] In some cases, the combined delay of unit dispatch and data transmission can be 1-2 hrs, which may be longer than the trucks' dwell times at various locations.

Considering the similarity between CWHT transportation behavior and human travel behavior, we begin this review with studies on individual mobility prediction and investigate modeling methods that can be applied. So far, many studies have analyzed individual mobility patterns (Cheng et al., 2021; Goulet-Langlois et al., 2018; Ma et al., 2020; Ma et al., 2013; Zhang et al., 2019; Zhao et al., 2018a) and developed various individual mobility models (Chen et al., 2022; Ma & Zhang, 2022). These models can generally be divided into two categories: statistical models and deep learning-based models.

Statistical models have been widely used in individual mobility prediction. Many researchers have proposed prediction models based on the Markov chains. which predicts the next location based on historical location information (Calabrese et al., 2010; Gambs et al., 2012; Gidófalvi & Dong, 2012; Lu et al., 2013; Neto et al., 2018). Monreale et al. (2009) used a T-pattern decision tree to learn from trajectory patterns and to predict the next location by finding the best matching path in the tree. Mathew et al. (2012) proposed a method for predicting human mobility based on the hidden Markov model utilizing historical features of user movement. Alvarez-Garcia et al. (2010) proposed a method for predicting human mobility based on the hidden Markov model utilizing historical features of user movement. Given the structural similarities between travel behavior and natural language, some natural language processing models have also been used for travel sequence prediction. For example, Hsieh et al. (2015) proposed a new time-aware language model called T-Gram, which uses location sequence data to predict human mobility. Zhao et al. (2018b) proposed a new model based on the Bayesian N-Gram model used in language modeling, which estimates the probability distribution of the next arrival location using previous travel information. Based on statistical prediction models, transition probabilities between distinct movement patterns are extracted from training data and used for location prediction. However, these models may struggle to completely account for the different factors that drive individual travel behavior and may be constrained when studying the periodicity in travel patterns.

Given the rapid development of deep learning models in recent years, various individual mobility prediction methods based on deep learning have emerged. Recurrent neural networks and their variants have been widely used for next-location prediction (Al-Molegi et al., 2016; Guan et al., 2023; Liu et al., 2019; Yao et al., 2023; Zhang et al., 2022). Such models can capture high-order spatiotemporal dependencies and periodic features, demonstrating better performance than statistical models. Many studies have also introduced attention mechanisms to learn the spatiotemporal dependencies between historical travel sequences (Feng et al., 2018; Li et al., 2020; Rossi et al., 2020). The individual mobility prediction models based on deep learning have the same modeling structure. They project the input trajectory information into a semantic space, transforming it into a feature vector containing a series of spatiotemporal information. Through sequence modeling, the dependency relationships of the travel patterns corresponding to these feature vectors are captured, and the final prediction result is output by the prediction module.

Although deep learning models have good prediction performance, the interpretability of these models is not as good as that of statistical models, which makes them difficult to understand by policymakers and reduces their applicability.

To address the limitations of statistical methods and deep learning methods, researchers have developed an innovative hidden Markov model, IOHMM, to integrate the two types of methods (Bengio & Frasconi, 1994). The advantage of IOHMM lies in its ability to not only enhance the model prediction performance by capturing rich contextual information but also to generate predictions based on diverse patterns. The model has also been applied in the study of individual mobility patterns. For example, researchers have used IOHMM to predict the next location and time of arrival of travelers based on vehicle GPS data (Hu et al., 2015), mobile signaling data (Yin et al., 2018), ride-sharing trip data (Zhang et al., 2021), and transit smart card data (Mo et al., 2022). Inspired by the similarity between the transportation activity patterns of CWHTs and human mobility patterns, this study adopts the IOHMM to predict the next transportation activity of CWHTs, including the destination and duration.

## 3. Data and Methodology

### 3.1. Description of the Dataset

The trajectory data of CWHTs in Chengdu, China were analyzed and used to demonstrate the proposed model. The trajectory data included the latitude, longitude, and time information of each trajectory point of CWHTs. We used the trajectory data of 300 CWHTs for a total of 3 months from September 2022 to November 2022 to develop the prediction model. To ensure that there is enough historical activity data, these trucks were randomly selected from all the trucks that made trips for at least 40 days during the study period. In addition, we collected weather information from the National Weather Science Data Center.

The first task was to identify the status of the CWHTs. We categorized the status of the CWHTs into traveling and staying. The latter not only included the status of the truck being parked (fully stopped) but also included the statute of the truck traveling at a low speed in locations such as the construction site and the dumping site. However, identifying whether the truck is traveling or staying is difficult. Some criteria must be set to distinguish the two statuses. One truck has many staying activities in a day, each of which corresponds to a destination. Because during the staying activity, the locations of the truck are still being collected. Thus, we can use $S_i$ to represent the set of consecutive trajectory points corresponding to the $i$-th staying activity.

$$S_i = \{(lat_1, lon_1, time_1), \ldots, (lat_j, lon_j, time_j), \ldots, (lat_n, lon_n, time_n)\},$$

where $(lat_1, lon_1, time_1)$, $(lat_j, lon_j, time_j)$, and $(lat_n, lon_n, time_n)$ correspond to the first, $j$-th, and last latitude, longitude, and time of the trajectory points in the corresponding staying activity.

According to the definition of stay points in Perez-Torres et al. (2016), the conditions that all trajectory points in staying activity $S_i$ should satisfy are

$$distance\left((lat_1, lon_1), (lat_j, lon_j)\right) \leq \theta_d, \forall 1 < j \leq n, \quad (1)$$

$$|time_1 - time_n| \geq \theta_t. \quad (2)$$

Equation (1) limits the size of a staying activity's geographic area, indicating that the distance between the first trajectory point and any other trajectory points must be smaller than $\theta_d$. Equation (2) determines the minimum time that a CWHT should spend at a staying activity. After determining the trajectory points that correspond to a staying activity, we use the centroid of the trajectory points to represent the location of the staying activity. Equation (3) and Equation (4) are used to calculate the longitude and latitude of the centroid.

$$c_{lat} = average(lat_j), j = 1,2,...,n \qquad (3)$$

$$c_{lon} = average(lon_j), j = 1,2,...,n \qquad (4)$$

Regulators typically use grids to divide cities into different squares for regional management. Thus, we divided Chengdu City into $62 \times 69$ squares with a side length of 2 km. The destinations of CWHTs are categorized into squares based on their centroids. We defined the location (or square) of the $i$-th staying activity in a day as $l_i$, the arrival time of the staying activity as $p_i$, and the departure time as $q_i$. Therefore, the staying activity $S_i$ can be represented as

$$S_i = (l_i, p_i, q_i).$$

For a particular CWHT, all the staying activities within one day can be represented as

$$SS = \{S_1, S_2 ..., S_m\},$$

where $m$ is the total number of staying activity in a day.

*3.2. Data Exploration*

Fig. 1 shows several measures related to the movements and operations of the CWHTs. Fig. 1 (a) illustrates the distribution of active days for the CWHTs, with an active day indicating that the truck generated at least one staying activity. Fig. 1 (b) displays the distribution of the number of trajectory records for each CWHT during the research period. The number of trajectory records for each CWHT predominantly ranges between 150 and 400. Fig. 1 (c) shows the duration of each CWHT spent at the staying activities, measured in minutes. The duration is usually shorter than 30 minutes, and seldom longer than 1 hour because the process of loading and unloading construction waste does not take a long time. In some cases, CWHT drivers may choose to take a rest near the construction site or dumping station, which leads to an extended stay duration. Fig. 1 (d) illustrates the distribution of the number of spatial units (squares) visited by each CWHT over the three months. Most CWHTs visited more than 20 squares within the three-month period. Fig. 1 (e) shows the distribution of time periods during which CWHTs generate staying activities, offering insights into the temporal distribution of their operational activities throughout the day. CWHTs generate staying activities throughout the day, with no consistent rest period. More CWHTs work in the time period from 6:00 to 15:00, whereas fewer CWHTs work in the time period from 21:00 to 5:00 the following day.

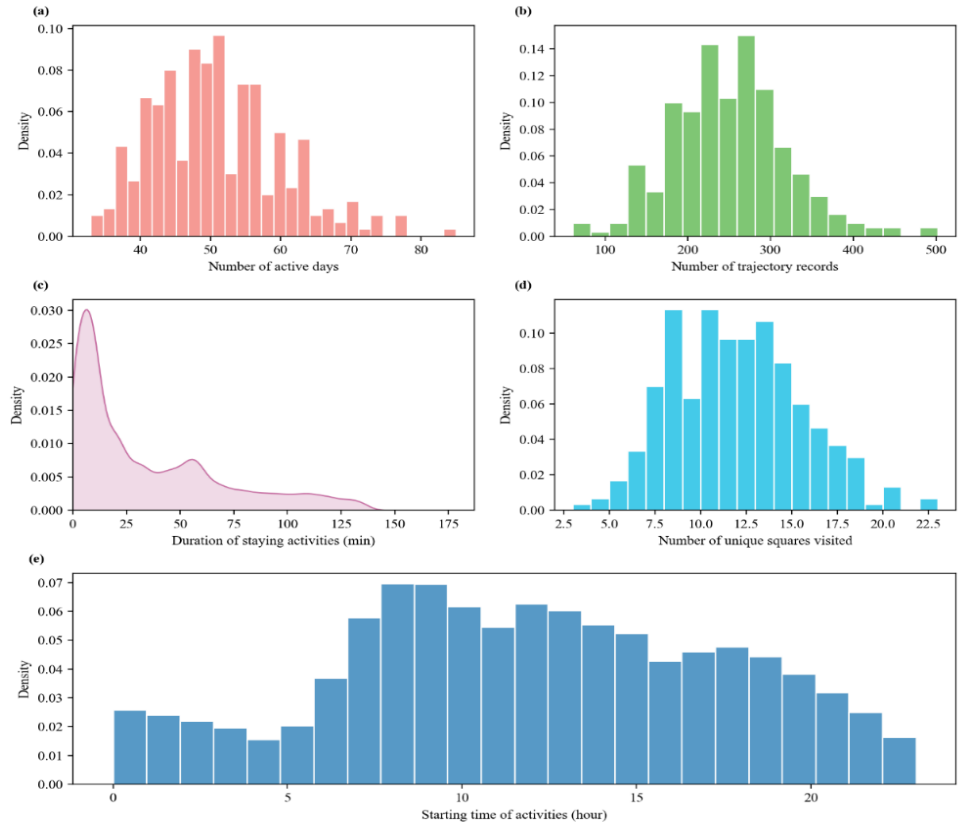

**Fig. 1  Characteristics of CWHTs' transportation activities**

### 3.3. Patterns of Transportation Activities

Regulators can track the possible activities of CWHTs based on their different transportation patterns. This information is key to predicting the destinations of CWHTs. First, the location types where CWHTs may stop must be explained. CWHTs, as an important carrier of building materials, have a wide range of destinations, including not only common places, such as parking lots, construction sites, and dumping stations, but also auto repair shops, gas stations, vehicle service stations, and others. Given the complexity of scenarios, we are currently unable to match the destinations with their corresponding types of locations. In the Chengdu region, CWHT transportation patterns exhibit two main types. Under the first transportation pattern, CWHTs depart from the parking lot every day to the construction site for loading activities. After transporting the construction waste to the dumping station, they return to the construction site for reloading and repeat the above transportation tasks until all the work is completed. CWHTs adopting this transportation pattern account for approximately 80%. In the second transportation pattern, an additional working location of a sand and gravel yard is incorporated into the previously described pattern. After leaving the dumping station, CWHTs may also go to the sand and gravel yard and then drive to the construction site until the task is accomplished. CWHTs adopting this transportation pattern account for approximately 20%. In the short term, the majority of CWHTs typically visit a single construction site each day. However, a small portion of CWHTs may travel to two or more construction sites, depending on the specific tasks assigned by the CWHT transportation company.

Although CWHT transportation patterns are similar to those of urban commuters, the destination types of CWHTs are more diverse and the starting time of transportation activities is more flexible. Furthermore, some factors that influence individual travel choices may not apply to CWHTs. For example, our findings reveal that factors such as weekdays and holidays have limited influence on CWHT behavior, suggesting the need to explore alternative factors.

### 3.4. Model Framework

In this study, we model each CWHT separately. The fundamental forecast interval is set at one day. The majority of CWHTs in Chengdu start their transportation work after 5:00, which can be seen in Fig. 1. Therefore, we set 5:00 as the starting and ending time of each day.

We defined all the staying activities for CWHT for a day; in particular, we define $t_i$ as the time spent to reach destination $l_{i+1}$ after driving out from $l_i$. For any $i = 1,2\ldots,m$, we have

$$t_i = p_i - q_{i-1}. \tag{5}$$

For the first trip of the day (often from a parking lot to a construction site), we define the departure time $q_0 = 5:00$ and the location $l = l_0$. When the CWHT leaves the parking site, the staying activity's location $l_i$ and the departure time $q_i$ are obtained. We aim to predict destination $l_{i+1}$ and the duration $t_{i+1}$ of the next transportation activity by using the historical transportation information of CWHTs and other related information. The model diagram is shown in Fig. 2.

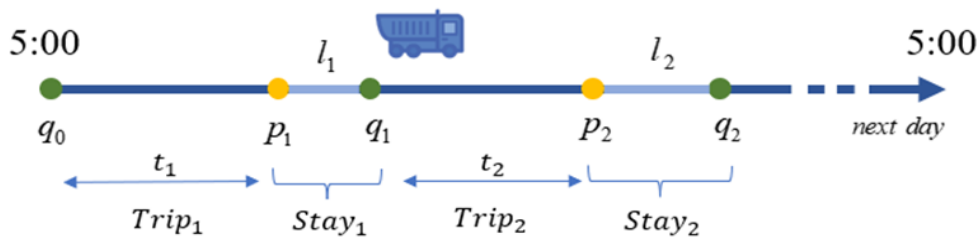

**Fig. 2 Modeling diagram for the daily transportation activities of CWHTs**

### 3.5. Input-Output Hidden Markov Model

IOHMM is an improved model based on HMM. HMM has had widespread application in trajectory prediction and semantic recognition. However, HMM implies that various observation states are exclusively related to their current hidden states. When the prediction scenarios are complicated and influenced by several elements, such as human mobility prediction, this model usually has poor performance. IOHMM can capture the changes in transition and emission probabilities that may occur through contextual information that varies over time, thereby adapting to more complex prediction scenarios. The architecture of IOHMM is well described in Bengio and Frasconi (1994). The framework of IOHMM for transportation activities prediction is shown in Fig. 3. IOHMM consists of three layers: the hidden layer, the input layer, and the output layer. The information in the input layer is the observed contextual variable $z_i$ (such as working hours, weather conditions, historical transportation information, etc.). The value of the input variable $z_i$ must be known prior to a transition. The hidden layer is mostly composed of latent categorical variables $h_i$ that represent the hidden state at time $i$. The output layer provides the observed variable $o_i$, which includes the destination $l_i$ and the duration $t_i$ of this transportation activity. $\lambda_{in}$, $\lambda_{tr}$, and $\lambda_{em}$ are the parameters of the initial, transition, and emission probability, respectively.

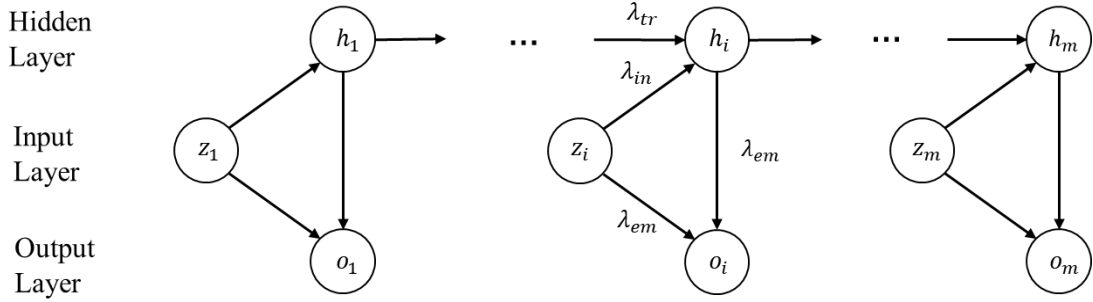

**Fig. 3 The framework of IOHMM for the prediction of transportation activities**

The model mainly includes three independent models: the initial model, the transition model, and the emission model. The initial model is used to calculate the initial state probability, $\pi_u = P(h_1 = u|z_i; \lambda_{in})$, $u \in H$, where $H$ is the state space, and it represents the distribution of the first activity type. The transition model is used to calculate the state transition probability, $\varphi_{uv,i} = P(h_i = v|h_{i-1} = u, |h_{i-1} = u, z_i; \lambda_{tr})$, where it represents the probability of state $u$ transitioning to state $v$. The emission model is used to calculate the emission probability, $\delta_{u,i} = P(o_i|h_i = u, z_i; \lambda_{em})$, where it represents the probability of the activity destination and activity duration under state $u$. Under this model, the likelihood of the data sequence can be represented using the following formula:

$$L(\lambda) = \sum_{h_1,\ldots,h_m} P(h_1|z_1; \lambda_{in}) \cdot \prod_{i=2}^{m} P(h_i|h_{i-1}, z_i; \lambda_{tr}) \cdot \prod_{i=1}^{m} P(o_i|h_{i-1}, z_i; \lambda_{em}), \tag{6}$$

where $\lambda = \{\lambda_{in}, \lambda_{tr}, \lambda_{em}\}$.

The parameters are estimated using the expectation–maximization (EM) algorithm, which is widely used in IOHMM parameter estimation. We used the code developed by Yin et al. (2018). The EM algorithm consists of two steps, as follows:

*E-step*: On the basis of the observed data and parameters estimated in the previous step, the expected value of the likelihood of the complete data sequence is calculated. We define the estimated parameters after $k - 1$ iterations of the M-step as $\lambda^{(k-1)}$, where

$$\lambda^{(k-1)} = \left(\pi_u^{(k-1)}, \varphi_{uv,i}^{(k-1)}, \delta_{u,i}^{(k-1)}\right).$$

Then, we can use $\lambda^{(k-1)}$ to calculate the forward vector $\alpha_{u,i}^{(k)}$ and the backward vector $\beta_{u,i}^{(k)}$ as follows:

$$\alpha_{u,i}^{(k)} = P(o_{1:i}, h_i = u|z_{1:i}) = \delta_{u,i}^{(k-1)} \sum_{u \in H} \varphi_{uv,i}^{(k-1)} \cdot \alpha_{u,i-1}^{(k)}, \tag{7}$$

$$\beta_{u,i}^{(k)} = P(o_{i+1:m}, |h_i = u, z_{i:m}) = \sum_{u \in H} \varphi_{u,i}^{(k-1)} \cdot \beta_{u,i+1}^{(k)} \cdot \delta_{u,i+1}^{(k-1)}, \tag{8}$$

where $\alpha_{u,1}^{(k)} = \pi_u^{(k-1)} \delta_{u,i}^{(k-1)}$, $\beta_{u,m}^{(k)} = 1$, The subscript $1:i$ indicates the set of variables corresponding to indices 1 to $i$.

Then, we can calculate the posterior state probability and posterior transition probability as follows:

$$\gamma_{u,i}^{(k)} = P(h_i = u | o_{1:m}, z_{1:m}) = \alpha_{u,i}^{(k)} \cdot \beta_{u,i}^{(k)} / \sum_{u \in H} \alpha_{u,m}^{(k)}, \tag{9}$$

$$\zeta_{uv,i}^{(k)} = P(h_i = v, h_{i-1} = u | o_{1:m}, u_{1:m}) = \varphi_{uv,i}^{(k-1)} \cdot \alpha_{u,i-1}^{(k)} \cdot \beta_{u,i}^{(k)} \cdot \delta_{u,i}^{(k-1)} / \sum_{u \in H} \alpha_{u,m}^{(k)}. \tag{10}$$

Finally, we obtained the parameters $\alpha_{u,i}^{(k)}, \beta_{u,i}^{(k)}, \gamma_{u,i}^{(k)}$ and $\zeta_{uv,i}^{(k)}$ in the E-step, which are then used in the M-step to estimate the maximum likelihood of the data.

*M-step*: Model parameters are updated by maximizing the expected complete data likelihood.

$$\begin{aligned} Q(\lambda, \lambda^{(k-1)}) = &\sum_{u=1} \gamma_{u,1}^{(k)} \log P(h_1 = u | z_1; \lambda_{in}) \\ &+ \sum_{i=2}^{m} \sum_{u} \sum_{v} \zeta_{uv,i}^{(k)} \log P(h_i = v | h_{i-1} = u, z_i; \lambda_{tr}) \\ &+ \sum_{i=1}^{m} \sum_{u} \gamma_{u,i}^{(k)} \log P(o_i | h_i = u, z_i; \lambda_{em}), \end{aligned} \tag{11}$$

where $\lambda^{(k)} = \operatorname{argmax}_\lambda Q(\lambda; \lambda^{(k-1)})$, and $k$ represents the iterations for the EM algorithm.

### 3.6. Model Specification

*Contextual information*: We primarily select context variables $z_i$ from three dimensions, which include the cumulative activities of CWHTs in the past. As shown in Table 1, it mainly includes weather data, information about the most recent transportation activities, and historical work statistics. The inclusion of weather data as the context variable is primarily driven by the fact that weather conditions can significantly influence the transportation activities of CWHT. For instance, rainy and snowy weather can lead to slippery roads, which could potentially extend transportation durations and increase the risk of accidents. Importantly, all the aforementioned information is accessible and obtainable prior to the CWHT's next transportation activity, enabling its integration into the prediction model effectively. In addition, in contrast to individual mobility, the transportation tasks of CWHTs are centrally dispatched by the CWHT transportation company; thus, their transportation activities are not affected by circumstances, such as weekdays and holidays. To simplify the calculation, certain influencing factors require processing to ensure their appropriate inclusion and impact in the model. For instance, we employ binary variables to represent weather conditions.

**Table 1**

Summary of Contextual Variables $z_i$.

| Variable | Type | Variable | Type |
| --- | --- | --- | --- |
| Sunny | Binary | Starting hour of the first trip on the previous day | Integer |
| Rainy | Binary | Starting hour of the last trip on the previous day | Integer |
| Cloudy | Binary | Number of consecutive days without work | Integer |
| Foggy | Binary | Number of trips on the previous day | Integer |
| Duration of the last trip | Continuous | Duration of the last staying activity | Continuous |

*Initial, transition, and emission models*: Given that the variables in the hidden layer are discrete, a multinomial logistic regression model is used in the initial probability model and transition probability model. In contrast to HMM, the initial and transition probabilities in IOHMM are determined by the context variables $z_i$. In particular, we have

$$P(h_1 = u | z_1; \lambda_{in}) = \frac{exp(\lambda_{in,u} \cdot z_1)}{\sum_{k \in H} exp(\lambda_{in,k} \cdot z_1)}, \quad (12)$$

$$P(h_i = v | h_{i-1} = u, z_i; \lambda_{tr}) = \frac{exp(\lambda_{tr,u}^v \cdot z_i)}{\sum_{k \in H} exp(\lambda_{tr,u}^k \cdot z_i)}, \quad (13)$$

where $\lambda_{in,u}$ is the coefficient for state $u$ in the initial state function, and $\lambda_{tr,u}^v$ is the coefficient for the case that the current state is $u$ and the next state is $v$ in the state transition probability function.

Regarding the emission probability, the observed variable $o_i$ is composed of the destination $l_i$ and the duration $t_i$ of the current transportation activity. $l_i$ is a discrete random variable, whereas $t_i$ is continuous. For a hidden activity, we assume that these two variables are independent of each other. Thus, we can calculate them separately as follows:

$$P(o_i | h_i = u, z_i; \lambda_{em}) = P(l_i | h_i = u, z_i; \lambda_{eml}) \cdot P(t_i | h_i = u, z_i; \lambda_{emt}). \quad (14)$$

As the destination $l_i$ of the transportation activity is discrete, the same multinomial logistic regression model can be utilized as the output model. For continuous variables, such as duration $t_i$, we use a linear model as the output model to represent the effect of context variables on the variable. We suppose that the transportation duration of the CWHT follows a Gaussian distribution. Its mean is a function of the parameter variables, which can describe the effect of context variables on the transportation duration well. The formulas are as follows:

$$P(l_i = l | h_i = u, z_i; \lambda_{eml}) = \frac{exp(\theta_{\lambda_{eml,u,l}} \cdot z_i)}{\sum_{l \in L} exp(\theta_{\lambda_{eml,u,l}} \cdot z_i)}, \quad (15)$$

$$P(t_i | h_i = u, z_i; \lambda_{emt}) = \frac{1}{\sqrt{2\pi}\sigma_u} e^{-\frac{(t_i - \lambda_{emt,u} \cdot z_i)^2}{2\sigma_u^2}}, \quad (16)$$

where $\lambda_{eml,u,l}$ is the parameter of the emission probability function when the current state is $u$ and the activity destination is $l$, $\lambda_{emt,u}$, and $\sigma_u$ are the parameters and standard deviation of the emission probability function when the current state is $u$. $L$ is the set of all destinations that the CWHT has been to. To simplify the model calculation, we make the assumption that the variance of activity duration remains constant for a specific type of transportation activity, based on the underlying characteristics of CWHT operations. The conditional independence assumption allows us to model the joint probability of $l_i$ and $t_i$ through hidden activities, making it feasible to estimate both variables simultaneously (Yin et al., 2018).

Prior to model training, it is necessary to ascertain the optimal number of hidden activity types for each CWHT. We employ a clustering approach, K-Means algorithm, to cluster the CWHT transportation activity sequences and experiment with various numbers of clusters. The best result from these experiments is subsequently chosen as the number of hidden activities. We set the collection of alternative cluster numbers as $K = \{3, 4, 5, 6, 7, 8\}$. We then determine the optimal cluster number by calculating the silhouette coefficient for each clustering result. Fig. 4 displays the distribution of hidden state numbers observed among the 300 CWHTs involved in the clustering process. As depicted in Fig. 4, the majority of CWHTs exhibit a hidden state number of 3, with a decline in proportion as the number of hidden states increases.

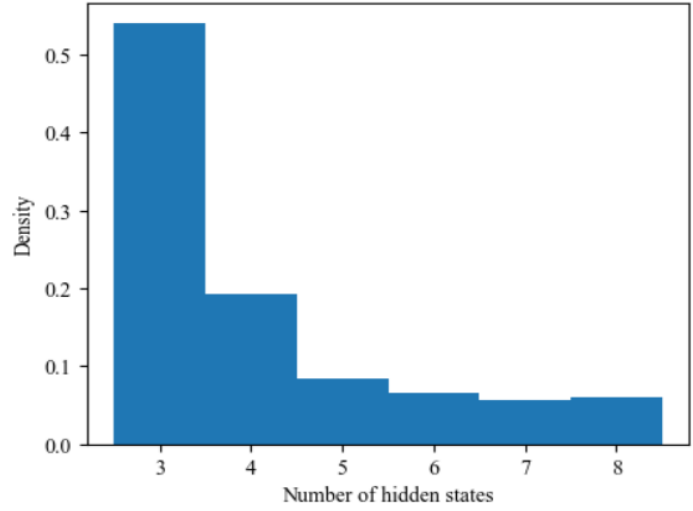

Fig. 4 Distribution of the number of hidden states

*3.7. Prediction Formulation*

We can use the trained IOHMM and transportation activities of a CWHT of the current day to predict the duration of the next transportation activity, as well as the destination. To achieve this goal, we must calculate the values of $P(o_{i+1}|o_{1:i}, z_{1:i+1})$. As $o_{1:i}, z_{1:i+1}$ are already observed, we can use the following formula to obtain this probability:

$$P(o_{i+1}|o_{1:i}, z_{1:i+1}) = \sum_{u \in H} P(o_{i+1}|h_{i+1} = u, z_i) \cdot P(h_{i+1} = u|o_{1:i}, z_{1:i+1}). \tag{17}$$

The first term on the right-hand side of the equation is the emission probability, whereas the second term requires further computation:

$$P(h_{i+1} = u|o_{1:i}, z_{1:i+1}) = \frac{P(h_{i+1} = u, o_{1:i}|z_{1:i+1})}{\sum_{v \in H} P(h_{i+1} = v, o_{1:i}|z_{1:i+1})}, \tag{18}$$

where:

$$P(h_{i+1} = u, o_{1:i}|z_{1:i+1}) = \sum_{u \in H} P(h_{i+1} = v|h_i = u, z_{i+1}) \cdot P(h_i = u, o_{1:i}|z_{1:i}). \tag{19}$$

The first term on the right-hand side of Equation (19) is the transition probability, and the second term is the forward probability, which can be calculated using the forward algorithm in HMM. We can calculate the forward probability, transition probability, and emission probability using the trained IOHMM. Thus, given the observed information $o_{i+1}$ and $z_{1:i+1}$, we can calculate the probability distribution of $o_{i+1}$, which includes the destination $l_{i+1}$ and the duration $t_{i+1}$ of the next transportation activity. The prediction model could generate the probability of all the potential destinations and time of arrival. The one, either location or the time of arrival, with the highest probability is selected as the prediction result.

**4.Results**

*4.1. Prediction of the Next Transportation Activity*

In this section, we demonstrate the use of IOHMM to construct a model for 300 participating CWHTs to predict the next transportation activity. To facilitate model training and evaluation, we partition the trajectory sequence of each CWHT into two distinct sets: a training set and a testing set. The training set consists of the mobile sequences of the first 70% of active days, whereas the remaining active days are used for testing. A model is trained using IOHMM for each CWHT based on its historical trajectory sequence.

The proposed model is designed to predict two variables: the destination $l_{i+1}$ and the duration $t_{i+1}$ of the next transportation activity. The former is discrete, whereas the latter is continuous. Therefore, two distinct indicators are used to assess the performance of the proposed prediction model. For destination $l$, we use the percentage of correctly predicted destinations as the prediction accuracy. For duration $t$, we use $R^2$ as the evaluation index. $R^2$ is often used to measure the fitting degree of regression, and its value is usually between 0 and 1. During the testing process, we calculate the prediction accuracy of each vehicle so that we can analyze the accuracy distribution of all CWHTs in the testing dataset to provide a more detailed evaluation of the prediction model.

To validate the performance of IOHMM, we must compare it with baseline models. We refer to the baseline models chosen by Mo et al. (2022) and compare them to two other types of models. The first type of model includes linear regression (LR) and Markov chain (MC), which are used to predict $l_{i+1}$ and $t_{i+1}$, respectively.

The calculation formula for the LR model is shown in Equation (20).

$$t_i = \beta_0 + \beta * z_i + \epsilon, \tag{20}$$

where $\beta_0$ is the intercept, $\beta$ is the vector of parameters to be estimated, and $\epsilon$ is the error term.

The formulas for the MC model are shown in Equations (21) and (22).

$$P(l_1) = \frac{C(l_1) + \alpha/|L|}{V + \alpha}, \tag{21}$$

$$P(l_i \mid l_{i-1}) = \frac{C(l_{i-1}, l_i) + \alpha/|L|}{\sum_{l_i \in L} C(l_{i-1}, l_i) + \alpha} (\forall t > 2), \tag{22}$$

where $C(l_1)$ is a counting function used to count the number of times the first transportation activity ends at $l_1$ in a day. Similarly, $C(l_{i-1}, l_i)$ counts the number of times the vehicle's transportation activity ends at $l_{i-1}$ and the next transportation activity ends at $l_i$ on the same day.

The second model is long short-term memory (LSTM). LSTM models have been widely used in time-series forecasting tasks due to their exceptional ability to capture and leverage long-term dependencies, enabling accurate predictions in various domains such as finance, weather forecasting, and stock market analysis. LSTM models can be effectively employed for predicting both continuous variables and discrete variables, making them suitable for a wide range of prediction tasks. We trained two independent models for each CWHT to predict $t_{i+1}$ and $l_{i+1}$, respectively. We tested multiple combinations of hyperparameters on the validation dataset, which consisted of 30% of the training data, to identify the optimal configuration for the models. The hyperparameters with the highest $R^2$ for the duration prediction and the highest percentage of correct predictions for the destination prediction in the validation data were selected as the final models' configuration. Specifically, for each CWHT, the LSTM model was configured with a single layer consisting of 50 units. We set the dropout rate to 0.2, the batch size to 64, and the number of training epochs to 120. We employed the Adam optimizer, a popular optimization algorithm known for its effectiveness in training neural networks, to optimize the model. Additionally, we utilized the mean squared error (MSE) as the loss function to measure the difference between predicted and actual values.

*4.2. Prediction Performance*

The IOHMM model is trained using a computer equipped with an AMD Ryzen 5 5600X CPU and it took approximately 1745.98 seconds to complete the training for 300 CWHTs. We evaluated the overall prediction performance by analyzing the distribution statistics of $R^2$ and the percentage of correct predictions. In summary, IOHMM performs much better in predicting both the duration of transportation activities and the location of destinations of the CWHTs. Specifically, in the duration prediction task, IOHMM outperformed LSTM and LR significantly. As shown in Fig. 5, the average $R^2$ of IOHMM, LSTM, and LR are 69.4%, 36.3%, and 32.5%, respectively. The majority of the predictions made by IOHMM have $R^2$ of higher than 60% while the majority of the predictions made by LSTM and LR have $R^2$ of lower than 60%.

In addition, we conducted a statistical analysis of the prediction error of the three duration prediction models. The distribution of absolute prediction errors for duration prediction is depicted in Fig. 6. As can be seen, the proportion of prediction errors for all models peaks at the category of lower than 0.5 hours and gradually decreases as the error increases. As anticipated, IOHMM outperforms the other two models with over 50% of the prediction errors lower than 0.5 hours.

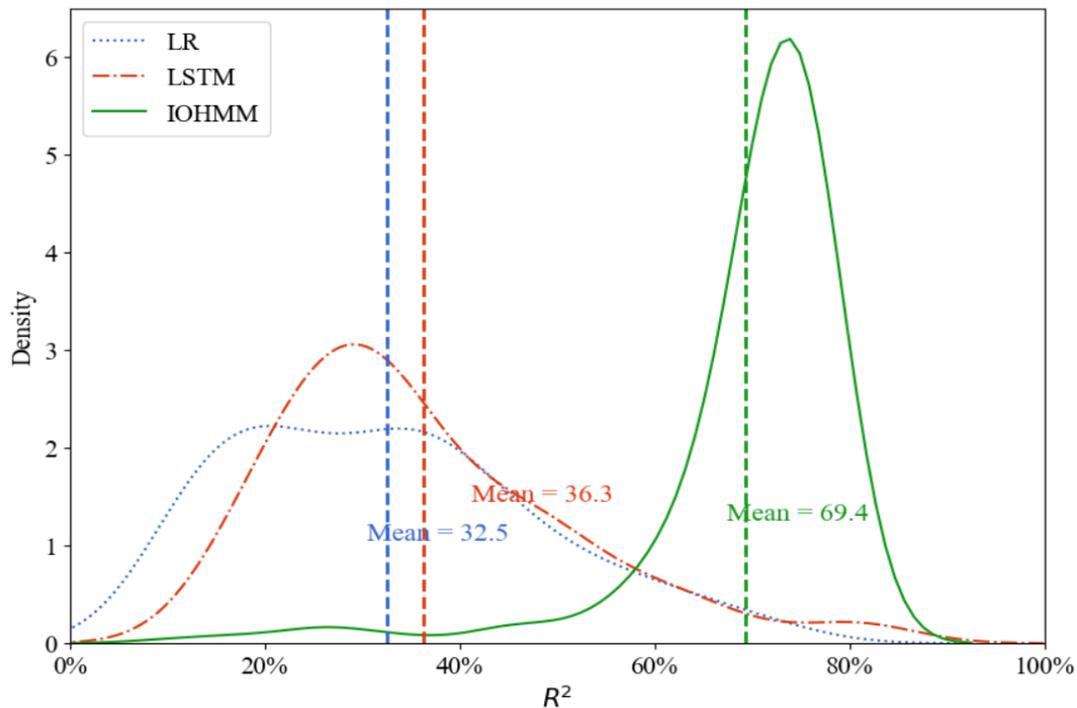

**Fig. 5 Accuracy distribution for transportation activity duration prediction**

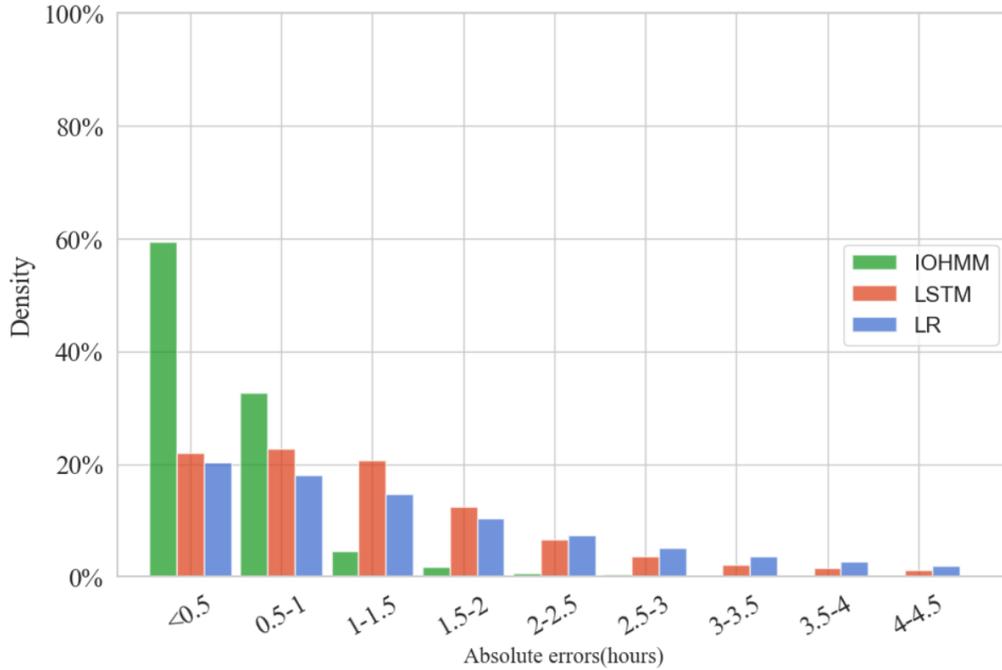

**Fig. 6 Distribution of prediction errors for transportation activities' duration**

In the destination prediction task, IOHMM is comparable to LSTM and outperforms MC. As depicted in Fig. 7, IOHMM still has the highest prediction accuracy (64.4%), followed by LSTM (60.9%) and MC (51.5%). It should also be noted that a higher proportion of the predictions made by IOHMM have a prediction accuracy of over 80%.

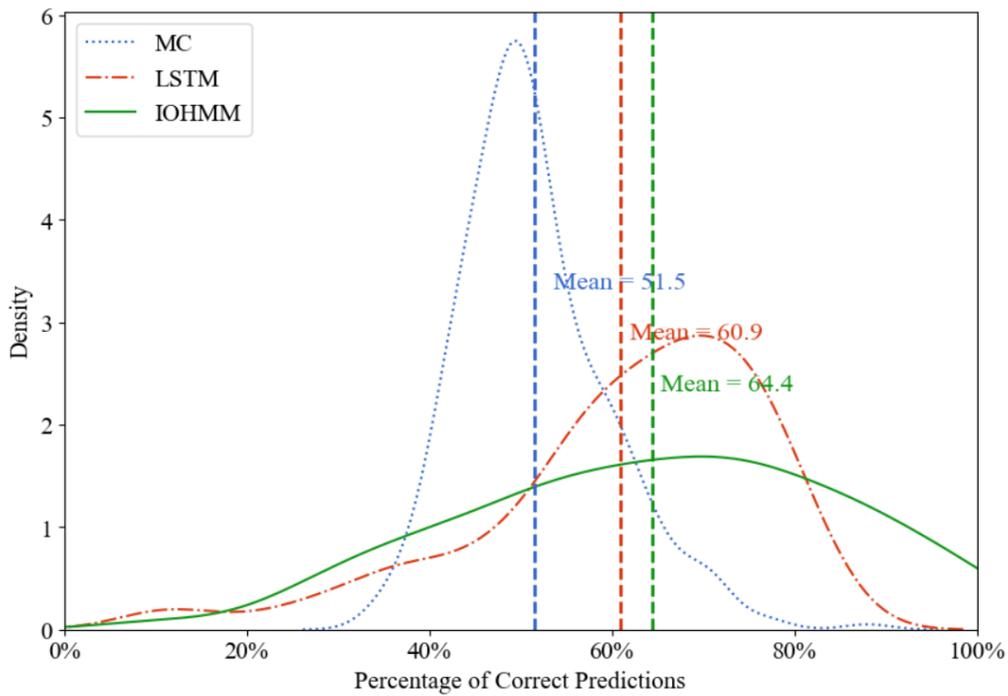

**Fig. 7 Accuracy distribution for destination prediction**

### 4.3. Factors Affecting Transportation Activity Prediction Accuracy

In this section, we evaluate the effect of different factors on the performance of the proposed IOHMM. These factors include the frequency and regularity of CWHTs' transportation activities, as well as CWHTs' attributes. To quantify the regularity of CWHTs' transportation activities, we utilized the standard deviation as a measure, where a high standard deviation indicates high variability and low regularity. Ultimately, seven different parameters are selected: the number of active days, the average number of daily transportation activities, the standard deviation of the number of daily transportation activities, the standard deviation of the duration of the first transportation activity of the day, the number of staying activities of all days, number of hidden states, and proportion of days with nighttime activities. The scatter plots of the prediction accuracy against different factors are shown in Fig. 8.

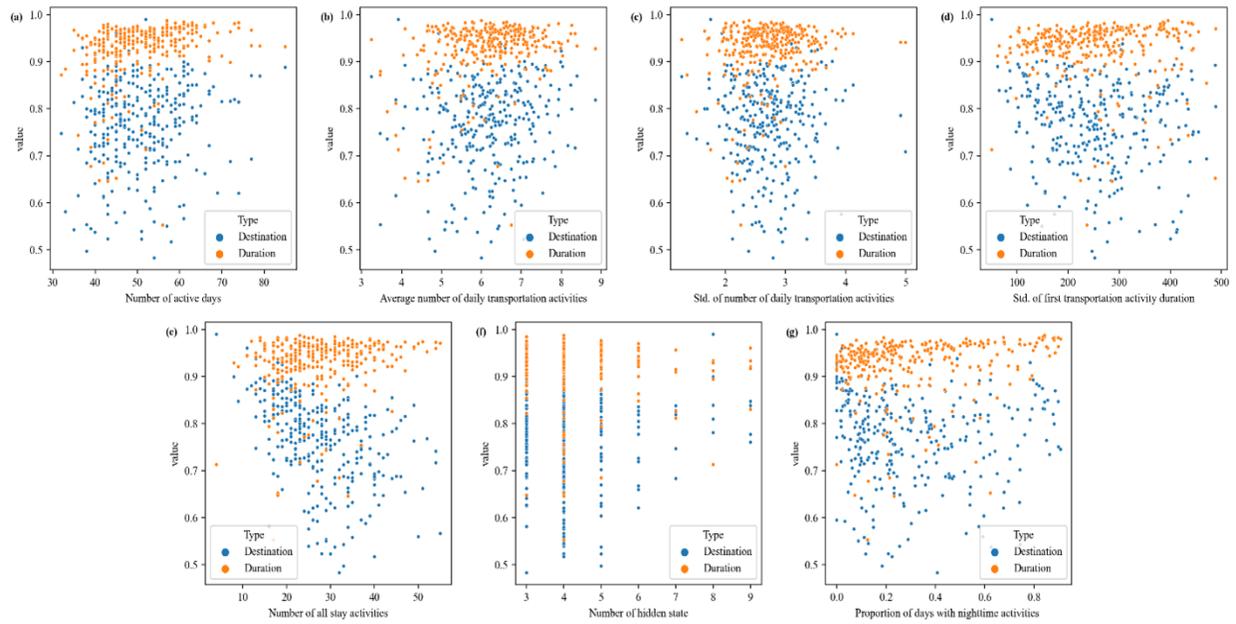

**Fig. 8 Scatter plot of $R^2$/Accuracy against different factors**

To investigate the effect of numerous factors on prediction performance, we use the two indicators, accuracy of destination prediction and $R^2$ of duration prediction, as dependent variables, and these influencing factors as independent variables to construct linear regression models. The results are shown in Table 2.

**Table 2**

Factors Influencing Predictability of CWHTs' Transportation Activities.

| Variable | Coefficient | |
| --- | --- | --- |
| | Destination | Duration |
| Intercept | 0.7145** | 0.6262 ** |
| Number of active days | 0.0018 ** | 0.0028 ** |
| Average number of daily transportation activities | 0.0224 * | -0.0037 |
| Std. of number of daily transportation activities | -0.0204 | 0.0206 |

| Std. of duration of the first transportation activity | $-9.0652 \times 10^{-5}$ | $-8.1614 \times 10^{-5}$ |
| --- | --- | --- |
| Number of all staying activities | -0.0051 ** | -0.0003 |
| Proportion of days with nighttime activities | 0.0013 ** | 0.0015 ** |
| Number of hidden states | 0.0030 | -0.0030 |

Number of CWHTs: 300

**: *p-value* <0.01; *: *p-value* < 0.05

According to our research findings, the number of active days has a significantly positive correlation with the accuracy of both the destination and duration prediction. This observation implies that CWHTs with a higher number of active days are more likely to make accurate predictions for both destination and duration. Additionally, we found that the average number of daily transportation activities by CWHTs has a positive impact on destination prediction accuracy. These findings are intuitively comprehensible: CWHTs engaged in more frequent transportation activities assist the model in capturing their long-term transportation behavior more effectively, thereby enhancing prediction accuracy. Furthermore, we explored the influence of the proportion of nighttime work shifts on prediction accuracy, especially considering that many CWHTs operate on a two-shift work system. One plausible explanation for this observation is that CWHTs working during the night tend to have a more limited range of destinations, which contributes to their heightened predictability. Moreover, there exists a strong inverse relationship between the number of different staying activities of all days by CWHTs and destination prediction accuracy. In general, a smaller number of different staying activities indicates a simpler transportation pattern, resulting in improved predictive results. Interestingly, we discovered that parameters related to the frequency of CWHT transit were not significant factors in predicting both destination and duration.

## 5. Conclusions

In this paper, we focus on real-time prediction of CWHTs' transportation activities based on their GPS data. Such a prediction allows environmental law enforcement units to respond to potential infringement in a timely manner. We propose a method based on the IOHMM, which leverages contextual information and the historical transportation activities of the CWHT to predict the destination and duration of the next transportation activity. Our study employs a rich dataset of CWHT trajectories collected over a three-month period in Chengdu, China.

Our methodology involves several key steps. Initially, we identify and categorize staying activities, effectively dividing the trajectories into distinct spatial squares and subsequently generating a sequence of transportation activities for the CWHT based on the information obtained for the staying activities. A clustering algorithm is then applied to classify CWHT trajectory sequences into different patterns. On the basis of the clustering result, we train the prediction model using factors such as weather conditions and historical transportation activity statistics as input features. We compare the proposed model's performance to that of other prediction models to validate its performance. Based on the evaluation metrics, the proposed model demonstrates comparable destination prediction capability to LSTM, with an average accuracy of 64.4%. Furthermore, when it comes to predicting arrival time, our model outperforms LSTM with an average $R^2$ value of 69.4%. Notably, the proposed model also outperforms other baseline models in predicting both activity destination and duration.

In addition to evaluating model performance, we explore various factors that may influence the predictive accuracy of our CWHT prediction model. We find a positive correlation between the number of active days and the proportion of days with nighttime activities with model performance. Furthermore, our exploration of the destination prediction model highlights two additional influential factors: the number of daily transportation activities and the number of different staying activities of on all days. We observe a positive correlation between the number of daily transportation activities and destination prediction accuracy, indicating that CWHTs engaged in more frequent daily transportation activities yield more precise destination predictions. Conversely, a negative correlation is noted between the number of different staying activities of all days and destination prediction accuracy, suggesting that simpler transportation patterns, characterized by fewer historical locations, lead to enhanced destination prediction accuracy.

In sum, our study provides a predictive model for CWHT transportation activities, offering insights into the critical factors affecting prediction accuracy. This approach empowers environmental law enforcement agencies to proactively manage CWHTs and respond effectively to potential infringements, thereby contributing to improved urban environmental management.

Our study also has some limitations. One limitation is that we currently lack the ability to associate the destinations of CWHTs with specific location types due to the absence of available labels for these places. If we can obtain the types or labels of the locations, the predictive performance of our model and its interpretability could be further improved. Another limitation is that our prediction method relies on the trajectory data of CWHTs that is long enough. For CWHTs that have not generated a substantial amount of trajectory data, new methods must be developed for prediction.

**CRediT authorship contribution statement**

**Hongtai Yang**: Conceptualization, Methodology, Formal analysis, Writing – review and editing. **Boyi Lei:** Data curation, Formal analysis, Visualization, Writing – original draft. **Ke Han**: Conceptualization, Resources, Writing - review & editing. **Luna Liu**: Methodology, Validation, Writing - review & editing.

**Declaration of Competing Interest**

The authors declare that they have no known competing financial interests or personal relationships that could have appeared to influence the work reported in this paper.

**Declaration of Generative AI and AI-assisted technologies in the writing process**

During the preparation of this work, we used ChatGPT in order to polish the writing of this paper. After using this tool, we reviewed and edited the content as needed and take full responsibility for the content of the publication.

**Acknowledgments**

This study was funded by the Fundamental Research Funds for the Central Universities (no. 2682023ZTPY012), Guangdong Science and Technology Strategic Innovation Fund (the Guangdong-Hong Kong-Macau Joint Laboratory